\documentclass[conference]{IEEEtran}
\usepackage{cite}
\usepackage{amsmath,amssymb,amsfonts}
\usepackage{algorithmic}
\usepackage{graphicx}
\usepackage{textcomp}
\usepackage{xcolor}
\usepackage{booktabs}
\usepackage{fancyhdr}
\usepackage{etoolbox}

\fancypagestyle{firstpage}{
    \fancyhf{}
    \fancyhead[C]{\small A PREPRINT}
    \fancyfoot[C]{\small This work is a preprint and has not been peer-reviewed. It has been posted to share preliminary findings. The views expressed here are those of the authors and do not necessarily reflect the views or policies of ISRO SAC.  }

}
\def\BibTeX{{\rm B\kern-.05em{\sc i\kern-.025em b}\kern-.08em
    T\kern-.1667em\lower.7ex\hbox{E}\kern-.125emX}}
\begin{document}

\title{Towards Seamless Lunar Mosaics: Deep Radiometric Normalization for Cross-Sensor Orbital Imagery Using Chandrayaan-2 TMC Data\

}

\author{\IEEEauthorblockN{1\textsuperscript{st} Pratincha Singh}
\IEEEauthorblockA{\textit{Department of IOT and Intelligent Systems} \\
\textit{Manipal University Jaipur}\\
Jaipur,India \\
pratincha.229311065@muj.manipal.edu}
\and
\IEEEauthorblockN{2\textsuperscript{nd} Jai Gopal Singla}
\IEEEauthorblockA{\textit{Space Applications Centre} \\
\textit{Indian Space Research Organization}\\
Ahmedabad,India \\
jaisingla@sac.isro.gov.in}
\and
\IEEEauthorblockN{3\textsuperscript{rd} Prashant Hemrajani}
\IEEEauthorblockA{\textit{Department of IOT and Intelligent Systems} \\
\textit{Manipal University Jaipur}\\
Jaipur,India \\
prashant.hemrajani@manipal.edu}
\and
\IEEEauthorblockN{4\textsuperscript{th} Nitant Dube}
\IEEEauthorblockA{\textit{Space Applications Centre} \\
\textit{Indian Space Research Organization}\\
Ahmedabad, India \\
nitant@sac.isro.gov.in}
\and
\IEEEauthorblockN{5\textsuperscript{th} Amitabh }
\IEEEauthorblockA{\textit{Space Applications Centre} \\
\textit{Indian Space Research Organization}\\
Ahmedabad,India \\
amitabh@sac.isro.gov.in}
\and
\IEEEauthorblockN{6\textsuperscript{th} Hinal Patel}
\IEEEauthorblockA{hinalbpatel@gmail.com} 

}

\maketitle
\thispagestyle{firstpage}
\begin{abstract}
Radiometric inconsistencies remain a major challenge in generating seamless lunar mosaics from multi-mission orbital imagery due to variability in illumination geometry, sensor characteristics, and acquisition conditions. This paper presents a deep learning-based radiometric normalization framework for multi-mission lunar mosaics constructed primarily from ISRO's Chandrayaan-2 Terrain Mapping Camera (TMC) data, supplemented with auxiliary imagery from the SELENE (Kaguya) mission.

The proposed approach employs a conditional generative adversarial network (cGAN) comprising a U-Net-based generator and a PatchGAN discriminator to learn a nonlinear radiometric mapping from conventionally mosaicked lunar imagery to a photometrically consistent reference derived from LROC Wide Angle Camera (WAC) data. A patch-based training strategy with overlap-aware inference is adopted to enable scalable processing of large-area mosaics while preserving structural continuity across tile boundaries.

Quantitative evaluation using Structural Similarity Index (SSIM), Peak Signal-to-Noise Ratio (PSNR), and Root Mean Square Error (RMSE) demonstrates consistent improvements over traditional histogram-based normalization techniques. The proposed framework achieves enhanced tonal uniformity, reduced seam artifacts, and improved structural coherence across multi-source lunar datasets.

These results highlight the effectiveness of learning-based radiometric normalization for large-scale planetary mosaicking and demonstrate its potential for generating high-fidelity lunar surface maps from heterogeneous orbital imagery.
\end{abstract}


\section{Introduction}

Lunar surface mosaics play a crucial role in planetary science, enabling geological interpretation, surface characterization, and mission planning. Recent lunar missions have produced extensive orbital image datasets with complementary spatial coverage and resolution. Among these, ISRO’s Chandrayaan-2 Terrain Mapping Camera (TMC) provides high-resolution grayscale imagery that serves as a primary source for lunar cartographic products \cite{isrodata}. However, constructing large-scale mosaics using a single mission is often insufficient due to coverage gaps, illumination variability, and acquisition constraints. Consequently, integration of imagery from additional missions such as SELENE (Kaguya) becomes necessary to achieve comprehensive coverage \cite{jaxadata}.

Despite the availability of multi-mission datasets, generating seamless lunar mosaics remains a significant challenge due to radiometric inconsistencies. Variations in solar illumination angle, sensor response characteristics, exposure conditions, and imaging geometry introduce substantial brightness and contrast differences between adjacent image tiles. These discrepancies manifest as visible seam artifacts and tonal discontinuities in the final mosaics, adversely affecting both visual quality and scientific interpretability. Traditional techniques such as global histogram matching and block-based tone balancing attempt to address these issues with minimal parameter tuning; however, their robustness is limited when dealing with heterogeneous multi-source datasets \cite{jobson1997, rahman2004}.

\begin{figure}[t]
\centering
\includegraphics[width=0.95\columnwidth]{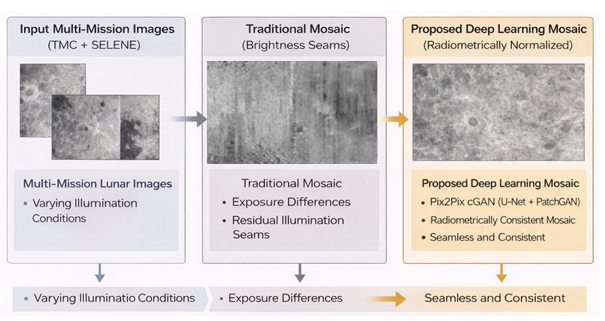}
\caption{Overview of the proposed deep learning framework for radiometric normalization of multi-mission lunar mosaics. The pipeline illustrates preprocessing, patch-based learning, and reconstruction stages for generating radiometrically consistent mosaics.}
\label{fig:overview}
\end{figure}

Recent advances in deep learning have enabled data-driven approaches for solving complex image normalization and translation tasks. In particular, conditional generative adversarial networks (cGANs) have demonstrated strong performance in image-to-image translation by learning nonlinear mappings between input and target domains \cite{isola2017}. Architectures such as U-Net facilitate effective feature preservation across multiple scales \cite{ronneberger2015}, while PatchGAN discriminators enforce local structural consistency. Although these approaches have been successfully applied in terrestrial imaging applications, their use in large-scale lunar mosaicking, particularly for radiometric normalization across multi-mission datasets, remains relatively unexplored.

The ability to generate radiometrically consistent mosaics has broader implications beyond visualization. High-quality mosaics are essential for accurate geomorphological analysis, surface change detection, and mission planning for future lunar exploration. Conventional methods often require extensive manual tuning and are sensitive to varying imaging conditions, limiting their scalability for large datasets. A learning-based approach can address these limitations by automatically adapting to variations in illumination and sensor characteristics.

In this work, we propose a deep learning-based radiometric normalization framework for seamless mosaicking of multi-mission lunar imagery. The proposed approach leverages Chandrayaan-2 TMC data as the primary dataset, supplemented with SELENE imagery, and utilizes a conditional GAN architecture to learn nonlinear radiometric transformations. A patch-based training strategy combined with overlap-aware inference enables scalable processing of large mosaics while preserving structural continuity.

The main contributions of this work are summarized as follows:

\begin{itemize}
    \item A deep learning-based radiometric normalization framework for multi-mission lunar mosaics, driven primarily by Chandrayaan-2 TMC data with auxiliary SELENE imagery.
    \item Application of a conditional GAN architecture with a U-Net generator and PatchGAN discriminator to learn nonlinear radiometric mappings for seam suppression and tonal harmonization.
    \item A patch-based training strategy with overlap-aware inference for scalable processing of large-area lunar mosaics.
    \item Comprehensive quantitative and qualitative evaluation demonstrating improved radiometric consistency over conventional histogram-based methods.
\end{itemize}

\section{Related Work}

Lunar image mosaicking has been widely studied using classical photogrammetric and image processing techniques. Early work primarily focused on geometric alignment and global radiometric balancing within single-mission datasets. For example, Wang et al. \cite{wang2011} proposed an automated pipeline for generating seamless global mosaics from Chang’E-1 CCD imagery, while Li et al. \cite{li2017} extended similar methodologies to Chang’E-2 data with improved tie-point extraction and optimization strategies. Although these approaches achieve reliable performance under controlled conditions, their applicability is limited when extended to multi-mission datasets characterized by significant radiometric variability.

Feature-based registration techniques such as SIFT \cite{lowe2004} and Laplacian-based matching \cite{zhang2020} have demonstrated strong performance in aligning lunar imagery. However, these methods primarily address geometric consistency and do not explicitly account for radiometric discrepancies across image tiles. As a result, mosaics generated using such techniques often exhibit visible seams and tonal discontinuities, particularly in regions with varying illumination conditions.

To address radiometric inconsistencies, several statistical and perceptual normalization techniques have been proposed. Methods based on histogram matching and Retinex theory \cite{jobson1997, rahman2004} attempt to standardize intensity distributions across images. While computationally efficient, these approaches are limited in handling nonlinear variations caused by changes in illumination geometry and sensor response. Photometric normalization strategies using reference datasets such as LROC WAC \cite{Sato1974} offer improved global consistency but require carefully calibrated reference mosaics and are less adaptable to heterogeneous multi-source data.

Recent advances in deep learning have introduced data-driven approaches for both image stitching and radiometric normalization. Deep homography estimation \cite{detone2016} and unsupervised image stitching frameworks such as UDIS++ \cite{nie2021} demonstrate the capability of neural networks to model complex transformations. In parallel, generative adversarial networks (GANs) have shown significant success in image-to-image translation tasks. The conditional GAN framework proposed by Isola et al. \cite{isola2017} enables learning of pixel-level mappings between domains, while histogram-aware GANs \cite{yang2019} incorporate distribution constraints to improve radiometric consistency. Additionally, Wasserstein GANs \cite{arjovsky2017} enhance training stability and convergence for generative models.

Despite these advancements, existing approaches predominantly focus on either geometric alignment or global intensity correction and are not specifically designed for large-scale radiometric normalization across multi-mission lunar datasets. The integration of heterogeneous imagery, such as Chandrayaan-2 TMC and SELENE data, introduces complex nonlinear radiometric variations that remain insufficiently addressed in current literature.

\begin{table*}[t]
\centering
\caption{Comparative Analysis of Existing Lunar Mosaicking and Radiometric Normalization Methods}
\label{tab:lit_review}
\begin{tabular}{p{2.5cm} p{3cm} p{2.5cm} p{2.5cm} p{3cm} p{3cm}}
\toprule
\textbf{Reference} & \textbf{Method} & \textbf{Dataset} & \textbf{Approach Type} & \textbf{Advantages} & \textbf{Limitations} \\
\midrule

Wang et al. \cite{wang2011} & Global mosaic generation & Chang’E-1 CCD & Classical & Automated pipeline, global consistency & Limited to single mission \\

Li et al. \cite{li2017} & Automated lunar mosaicking & Chang’E-2 CCD & Classical & Improved tie-point extraction & Poor multi-mission adaptability \\

Zhang et al. \cite{zhang2020} & Laplacian feature matching & Lunar imagery & Feature-based & Robust feature detection & No radiometric correction \\

Lowe \cite{lowe2004} & SIFT feature extraction & General images & Feature-based & Scale invariance & Not designed for radiometric normalization \\

Jobson et al. \cite{jobson1997} & Retinex-based enhancement & General imagery & Statistical & Enhances local contrast & Over-enhancement artifacts \\

Rahman et al. \cite{rahman2004} & Multiscale Retinex & General imagery & Statistical & Improved perceptual quality & Limited global consistency \\

Sato et al. \cite{Sato1974} & Photometric normalization & LROC WAC & Photometric & Physically grounded correction & Requires calibrated reference \\

DeTone et al. \cite{detone2016} & Deep homography estimation & General images & Deep Learning & Learns geometric transformations & No radiometric handling \\

Nie et al. \cite{nie2021} & UDIS++ stitching & Natural scenes & Deep Learning & Unsupervised learning & Limited planetary application \\

Isola et al. \cite{isola2017} & Conditional GAN (pix2pix) & General images & Deep Learning & Strong image translation capability & Requires paired data \\

Yang et al. \cite{yang2019} & Histogram-aware GAN & General images & Deep Learning & Distribution-aware learning & High computational cost \\

Arjovsky et al. \cite{arjovsky2017} & Wasserstein GAN & General images & Deep Learning & Stable training & Complex optimization \\

\textbf{Proposed Method} & cGAN-based normalization & TMC + SELENE & Deep Learning & Seam suppression, multi-mission consistency & Computational complexity \\

\bottomrule
\end{tabular}
\end{table*}

Table~\ref{tab:lit_review} summarizes the key characteristics of existing methods and highlights their limitations in handling multi-mission lunar mosaics. In contrast to prior work, the proposed approach leverages a conditional GAN framework to learn nonlinear radiometric transformations across heterogeneous datasets. By integrating patch-based training with overlap-aware inference, the method ensures scalability while preserving structural continuity, thereby addressing both local seam artifacts and global tonal inconsistencies.

\section{Methodology}

This section describes the proposed deep learning-based framework for radiometric normalization and seamless mosaicking of multi-mission lunar imagery. The framework is designed to address nonlinear radiometric inconsistencies across heterogeneous datasets through a combination of preprocessing, conditional generative modeling, and patch-based reconstruction.

\subsection{Overview of the Proposed Framework}

The overall pipeline of the proposed method is illustrated in Fig.~\ref{fig:overview}. The framework consists of three primary stages: (i) preprocessing and patch extraction, (ii) radiometric normalization using a conditional generative adversarial network (cGAN), and (iii) reconstruction of the normalized mosaic using overlap-aware inference.

Given an input mosaic $I$ composed of multiple tiles with radiometric inconsistencies, the objective is to learn a mapping function $G$ such that:

\begin{equation}
\hat{I} = G(I)
\end{equation}

where $\hat{I}$ represents the radiometrically normalized output aligned with a reference distribution derived from LROC WAC data.

\subsection{Preprocessing and Patch Extraction}

Prior to training, raw lunar images from Chandrayaan-2 TMC and SELENE datasets undergo preprocessing to enhance local contrast and remove non-informative regions. Contrast Limited Adaptive Histogram Equalization (CLAHE) is applied to improve local intensity variations while preventing over-amplification of noise.

Given the large spatial extent of lunar mosaics, the input images are divided into fixed-size patches of $512 \times 512$ or $1024 \times 1024$ pixels. Let $I \in \mathbb{R}^{H \times W}$ denote the input image. The patch extraction process can be expressed as:

\begin{equation}
P_{i,j} = I[x_i:x_i+s, \; y_j:y_j+s]
\end{equation}

where $s$ denotes the patch size and $(x_i, y_j)$ represent the spatial coordinates.

This patch-based strategy enables efficient training and improves generalization across varying radiometric conditions.

\subsection{Conditional GAN for Radiometric Normalization}

To model nonlinear radiometric transformations, a conditional generative adversarial network (cGAN) is employed. The cGAN framework consists of two components: a generator $G$ and a discriminator $D$.

\subsubsection{Generator Architecture}

The generator is based on a U-Net architecture \cite{ronneberger2015}, which enables multi-scale feature extraction and preservation of spatial details through skip connections. The encoder progressively downsamples the input patch to extract hierarchical features, while the decoder reconstructs the normalized output.

\subsubsection{Discriminator Architecture}

A PatchGAN discriminator \cite{isola2017} is used to evaluate local image patches instead of the entire image. This encourages high-frequency consistency and reduces local artifacts such as seams.

The discriminator learns to distinguish between real reference patches $I_{ref}$ and generated patches $\hat{I}$.

\subsection{Loss Function}

The overall objective function combines adversarial loss with pixel-level reconstruction loss:

\begin{equation}
\mathcal{L}_{cGAN}(G,D) = \mathbb{E}_{I,I_{ref}}[\log D(I, I_{ref})] + \mathbb{E}_{I}[\log (1 - D(I, G(I)))]
\end{equation}

To enforce structural similarity, an $L_1$ loss is incorporated:

\begin{equation}
\mathcal{L}_{L1} = \mathbb{E}_{I,I_{ref}} \left[ \| I_{ref} - G(I) \|_1 \right]
\end{equation}

The final objective becomes:

\begin{equation}
\mathcal{L}_{total} = \mathcal{L}_{cGAN} + \lambda \mathcal{L}_{L1}
\end{equation}

where $\lambda$ controls the trade-off between adversarial and reconstruction loss.

\subsection{Overlap-Aware Inference and Reconstruction}

During inference, patches are extracted with overlapping regions to minimize boundary artifacts. The final mosaic is reconstructed by blending overlapping patches using weighted averaging:

\begin{equation}
I(x,y) = \frac{\sum_{k} w_k(x,y) \cdot P_k(x,y)}{\sum_{k} w_k(x,y)}
\end{equation}

where $w_k(x,y)$ represents spatial weights for smooth blending.

\subsection{Hyperparameter Settings}

The model is trained using the Adam optimizer with a learning rate of $2 \times 10^{-4}$ and momentum parameters $\beta_1 = 0.5$, $\beta_2 = 0.999$. The batch size is set to 8 due to memory constraints, and training is conducted for 100–150 epochs.

The loss weighting parameter $\lambda$ is empirically set to 100 to balance adversarial and reconstruction losses. Patch sizes of $512 \times 512$ and $1024 \times 1024$ are used to analyze scalability and performance.

\section{Experimental Setup}

\subsection{Dataset Description}

\begin{table*}[t]
\centering
\caption{Lunar Datasets Used in This Study}
\label{tab:moon_data_summary}
\begin{tabular}{lccc}
\toprule
\textbf{Dataset} & \textbf{Source} & \textbf{Resolution} & \textbf{Usage} \\
\midrule
Chandrayaan-2 TMC & ISRO & 5 m & Primary dataset \\
SELENE (Kaguya) & JAXA & 8 m & Auxiliary dataset \\
LROC WAC & NASA/USGS & 100 m & Reference dataset \\
\bottomrule
\end{tabular}
\end{table*}

The proposed framework is evaluated using multi-mission lunar datasets. Chandrayaan-2 TMC imagery serves as the primary high-resolution dataset, while SELENE imagery is incorporated to enhance coverage. LROC WAC data is used as a radiometric reference for normalization.

\subsection{Implementation Details}

The framework is implemented in Python 3.11 \cite{pythonurl} using the PyTorch deep learning library \cite{paszke2019}. Preprocessing and visualization tasks are performed using QGIS \cite{qgisurl}. Training is conducted on an NVIDIA H100 NVL GPU.

\subsection{Training Configuration}

A patch-based training strategy is adopted to handle large-scale lunar imagery efficiently. Input mosaics are divided into overlapping patches to ensure continuity during reconstruction. Data augmentation techniques such as horizontal flipping and intensity scaling are applied to improve generalization.

Training is performed for 100–150 epochs, with model checkpoints saved periodically. Early stopping is applied based on validation loss to prevent overfitting.
\section{Results and Evaluation}

This section presents the experimental results of the proposed radiometric normalization framework applied to multi-mission lunar mosaics. The performance of the proposed deep learning framework for radiometric normalization was evaluated using both quantitative metrics and qualitative visual comparisons. The evaluation was performed by comparing the generated mosaics against the LROC WAC global mosaic, which served as the radiometrically consistent reference dataset. The model was trained for 150 epochs, and validation metrics were recorded at regular intervals.

The primary evaluation metrics used in this study were Peak Signal-to-Noise Ratio (PSNR) and Structural Similarity Index (SSIM), which measure radiometric similarity and structural consistency between generated images and the reference mosaic. Additionally, training convergence was monitored through generator and discriminator loss values.

\subsection{Experimental Setup}

Experiments were conducted using mosaicked lunar imagery derived primarily from Chandrayaan-2 Terrain Mapping Camera (TMC) data with supplementary SELENE observations. The initial mosaic was generated using conventional mosaicking techniques followed by preprocessing operations including contrast normalization and patch extraction.

The proposed conditional GAN model was trained on paired patches derived from the mosaicked input and a photometrically stable LROC Wide Angle Camera (WAC) reference mosaic. During inference, the trained model was applied to large mosaics using an overlap-based sliding window approach to ensure seamless reconstruction.

\subsection{Training Convergence}

During training, the adversarial learning process showed stable convergence. The generator loss decreased progressively as the network learned the mapping between the traditionally mosaicked input and the radiometrically normalized WAC reference. Simultaneously, the discriminator loss stabilized, indicating that the generated outputs became increasingly indistinguishable from the reference images.

\begin{table}[h]
\centering
\caption{Training loss values observed during model convergence}
\begin{tabular}{ccc}
\hline
Epoch & Generator Loss & Discriminator Loss \\
\hline
1   & 1.82 & 0.92 \\
50  & 0.76 & 0.54 \\
100 & 0.51 & 0.43 \\
125 & \textbf{0.44} & \textbf{0.39} \\
150 & 0.46 & 0.41 \\
\hline
\end{tabular}
\end{table}

The decreasing generator loss indicates improved reconstruction accuracy and enhanced tonal alignment with the WAC reference mosaic. The discriminator loss stabilizes around 0.4, suggesting balanced adversarial training without mode collapse.

\subsection{Training Progress Visualization}

\begin{figure}[h]
\centering
\includegraphics[width=\linewidth]{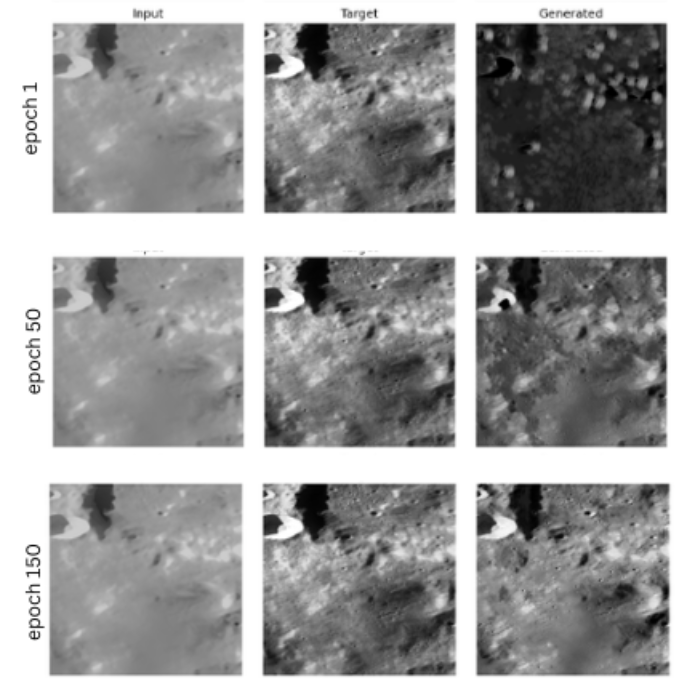}
\caption{Training progression showing generated mosaics at epoch 1, epoch 50, and epoch 150. The model gradually learns radiometric normalization across training iterations.}
\label{fig:training_progress}
\end{figure}

The progression of generated outputs during training further illustrates the effectiveness of the proposed approach. Figure~\ref{fig:training_progress} presents examples of generated mosaics at different epochs during training.

Early epochs produce outputs with noticeable tonal inconsistencies, whereas later epochs gradually learn the correct radiometric mapping between the input mosaic and the WAC reference. By epoch 150, the generator produces visually consistent mosaics with minimal seam artifacts.

\subsection{Quantitative Performance Evaluation}

To evaluate the effectiveness of the proposed approach, PSNR and SSIM values were computed for mosaics generated at different training epochs. These metrics quantify both pixel-level similarity and perceptual structural consistency.

\begin{table}[h]
\centering
\caption{Quantitative evaluation using PSNR and SSIM metrics}
\begin{tabular}{ccc}
\hline
Epoch & PSNR (dB) & SSIM \\
\hline
65  & 29.0442 & 0.9822 \\
75  & 28.7746 & 0.9811 \\
100 & 21.6414 & 0.9522 \\
150 & 29.6571 & 0.9818 \\
\textbf{125} & \textbf{30.5551} & \textbf{0.9876} \\
\hline
\end{tabular}
\end{table}

The highest performance was obtained at the 125th epoch, achieving a PSNR of 30.55 dB and an SSIM value of 0.9876. These values indicate a high level of radiometric and structural similarity with the reference WAC mosaic.

Compared to earlier training stages, the model demonstrates a clear improvement in tonal consistency and illumination correction. The temporary decrease in PSNR at epoch 100 reflects a typical instability phase in adversarial training before convergence.

\subsection{Visual Comparison of Mosaic Outputs}

\begin{figure}[h]
\centering
\includegraphics[width=\linewidth]{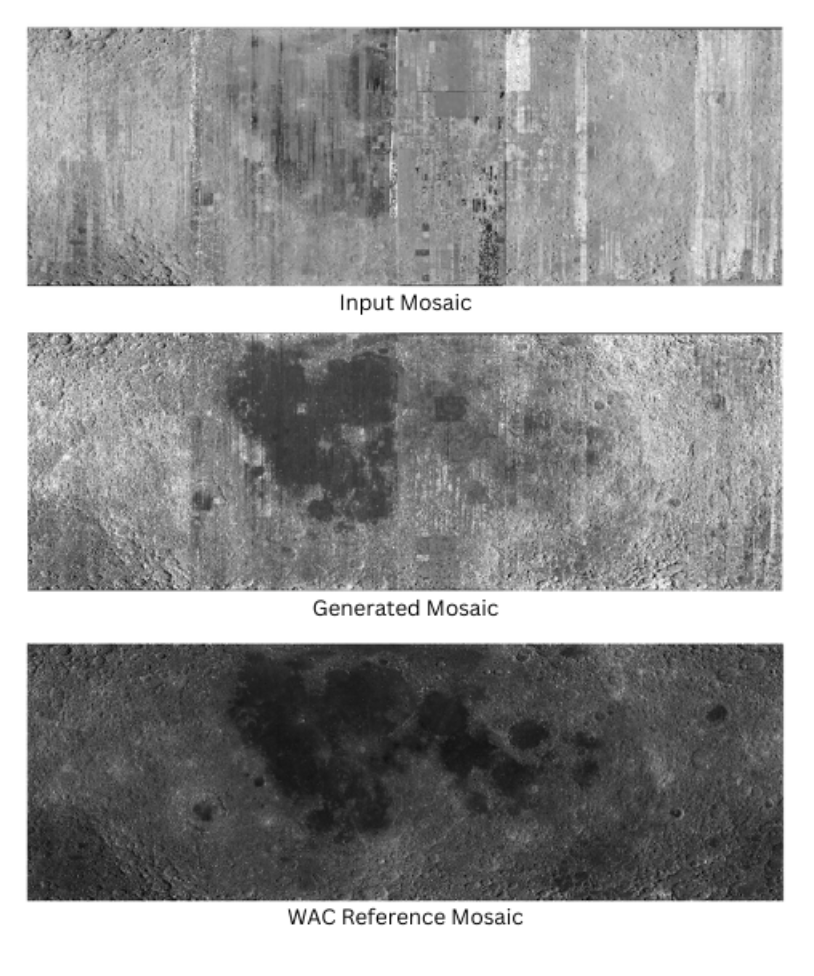}
\caption{Visual comparison between (a) mosiac generated using standard image processing approch, (b) mosiac generated using DL pipeline based on GANs, and (c) LROC WAC reference mosaic.}
\label{fig:mosaic_comparison}
\end{figure}

Figure~\ref{fig:mosaic_comparison} presents a visual comparison between the traditionally generated mosaic, the deep learning–normalized mosaic, and the LROC WAC reference mosaic. The baseline mosaic generated through classical histogram-based tone balancing exhibits noticeable brightness discontinuities across scene boundaries, particularly in regions where imagery from different sensors overlaps.

The proposed deep learning framework significantly reduces these discontinuities by learning a radiometric transformation that aligns the tonal distribution of the input mosaic with the WAC reference. As observed in the figure, the normalized mosaic demonstrates improved brightness uniformity and smoother transitions across adjacent tiles.

\subsection{Histogram Distribution Analysis}

\begin{figure}[h]
\centering
\includegraphics[width=\linewidth]{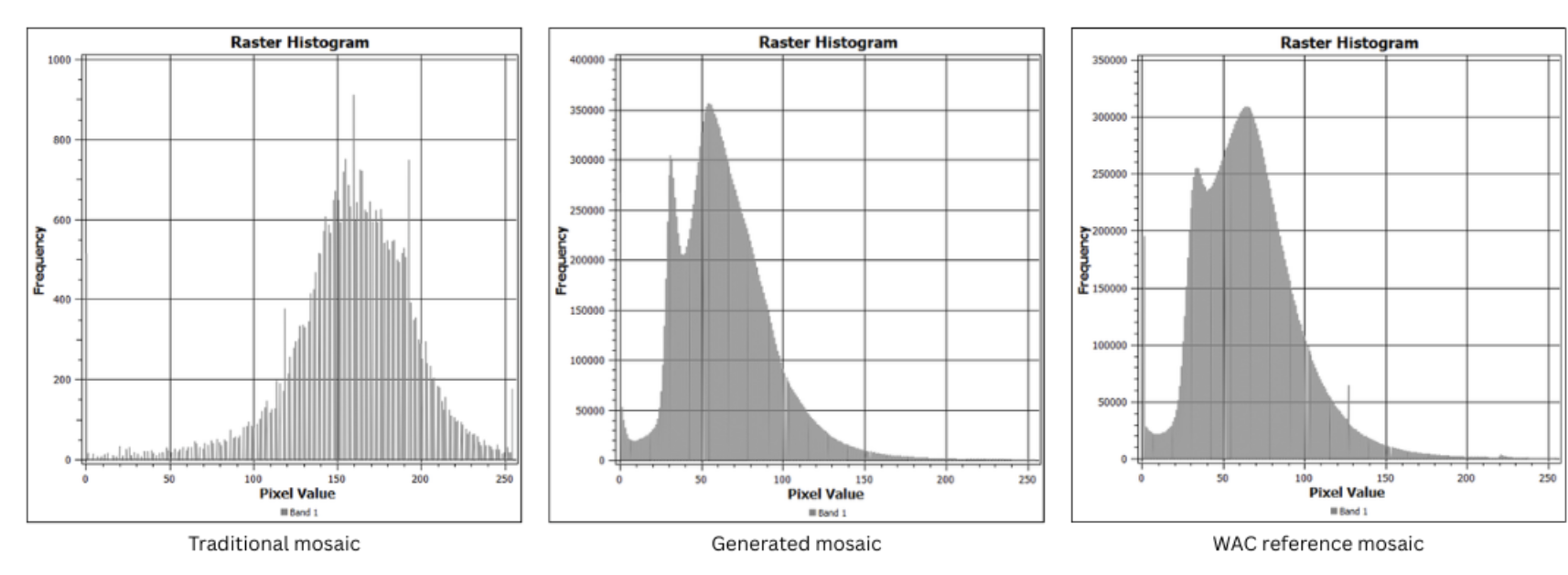}
\caption{Histogram comparison of the traditional mosaic, generated mosaic, and WAC reference mosaic showing improved radiometric alignment after normalization.}
\label{fig:histogram_comparison}
\end{figure}

To further evaluate radiometric consistency, histogram distributions of the mosaics were analyzed and compared with the reference WAC mosaic. Figure~\ref{fig:histogram_comparison} illustrates the intensity distribution of the traditional mosaic, the generated mosaic, and the reference mosaic.

The histogram of the traditional mosaic shows a broader distribution with irregular peaks, indicating inconsistent illumination and exposure differences across tiles. In contrast, the histogram of the generated mosaic closely aligns with that of the WAC reference mosaic, demonstrating that the proposed model successfully learns the underlying radiometric distribution.

\section{Discussion}

The experimental evaluation demonstrates that the proposed deep learning framework effectively improves radiometric consistency in multi-mission lunar mosaics. By learning a direct mapping between the traditionally mosaicked TMC–SELENE composite and the radiometrically stable LROC WAC reference, the network reduces tonal discontinuities that commonly appear at scene boundaries and sensor overlaps.

Unlike conventional histogram-based normalization methods, which apply global or block-wise intensity adjustments, the conditional GAN architecture learns spatially adaptive radiometric transformations. This enables the model to correct illumination differences while preserving fine-scale geological structures such as crater rims, ridges, and ejecta patterns. The high SSIM values observed in the experiments indicate that structural details of the lunar terrain are retained while improving tonal uniformity.

The histogram alignment analysis further confirms that the generated mosaics closely follow the radiometric distribution of the WAC reference dataset. This suggests that the proposed model captures not only pixel-level relationships but also broader statistical characteristics of lunar surface reflectance.

Another notable observation is the stability achieved after approximately 120 training epochs, beyond which performance improvements become marginal. This indicates that the network successfully converges to an optimal radiometric mapping without excessive overfitting.

Future work will focus on extending the framework toward higher spatial resolutions, particularly by incorporating full-resolution Chandrayaan-2 Terrain Mapping Camera (5 m) data and Orbiter High Resolution Camera (OHRC) imagery. Additional research may also explore multi-scale architectures and physics-informed learning strategies that incorporate photometric models of lunar illumination. These improvements could enable automated generation of globally consistent mosaics at significantly higher spatial resolutions.

Despite the promising results, several limitations should be acknowledged. First, the proposed model relies on the availability of a high-quality radiometric reference dataset. In this study, the LROC WAC global mosaic served as the reference domain; however, such reference datasets may not always be available for other planetary bodies or specific imaging conditions.

Second, the current training procedure operates on fixed-size image patches. While this approach enables efficient training and improves generalization, it may limit the model's ability to capture very large-scale illumination gradients that extend across multiple tiles. Although overlapping inference and Gaussian blending mitigate boundary artifacts, further improvements could be achieved through global-context learning mechanisms.

Another limitation arises from the adversarial training process itself. Generative adversarial networks are known to exhibit training instability under certain conditions. Although stable convergence was observed in this study, training requires careful hyperparameter tuning and monitoring to avoid mode collapse or oscillatory behavior.

Finally, the present framework focuses exclusively on radiometric normalization and does not address geometric misalignment or parallax effects. In practical large-scale mosaicking pipelines, geometric correction and radiometric harmonization must be integrated to produce fully seamless planetary maps.

\section{Conclusion}

This study presented a deep learning–based framework for improving radiometric consistency in large-scale lunar mosaics generated from multi-mission datasets. A conditional generative adversarial network composed of a U-Net generator and PatchGAN discriminator was trained to translate traditionally mosaicked TMC–SELENE imagery into an illumination-consistent representation aligned with the LROC WAC reference mosaic.

Quantitative evaluation using PSNR and SSIM metrics demonstrated that the proposed approach significantly improves tonal consistency while preserving structural characteristics of the lunar surface. The generated mosaics exhibit reduced brightness discontinuities and improved histogram alignment with the reference dataset, indicating effective correction of radiometric inconsistencies.

The results highlight the potential of deep generative models for planetary image processing, particularly in scenarios involving heterogeneous datasets acquired under varying illumination conditions. By combining traditional mosaicking pipelines with data-driven radiometric normalization, the proposed approach provides a scalable solution for generating high-quality lunar surface maps.

Future research will investigate integration with higher-resolution lunar datasets and explore hybrid frameworks that jointly address both geometric alignment and radiometric harmonization. Such developments could contribute to the automated generation of next-generation lunar mosaics for planetary science, mapping, and mission planning applications.

\end{document}